\documentclass{article}
\usepackage{spconf,amsmath,graphicx,hyperref,mathtools}
\usepackage{algorithm}
\usepackage{algorithmic}
\usepackage{xcolor}
\usepackage{amssymb}
\usepackage{booktabs} 
\usepackage{placeins}
\usepackage[table]{xcolor}
\usepackage[dvipsnames, svgnames, table]{xcolor}
\usepackage{multirow}
\usepackage{tabularx}

\usepackage{newfloat}
\usepackage{listings}
\usepackage{seqsplit}

\usepackage{caption} % 推荐先加载这个宏包，它提供了更强大的控制功能
\title{Enhancing Action and Ingredient Modeling for Semantically Grounded \\ Recipe Generation}

\name{Guoshan Liu$^{\star, \dagger}$, Bin Zhu$^{\dagger}$, Yian Li$^{\star}$,
Jingjing Chen$^{\star\ddag}$, Chong-Wah Ngo$^{\dagger}$, Yu-Gang Jiang$^{\star}$}
\address{%
\vspace{-0.6em} % ← 拉近“名字”和“机构”之间的间距；可在 -0.4em ~ -0.8em 间微调
\begin{tabular}{c}
$^{\star}$ Shanghai Key Lab of Intell. Info. Processing, School of Computer Science, Fudan University, China\\
$^{\dagger}$ Singapore Management University, Singapore\\
\{gsliu24,yali24\}@m.fudan.edu.cn, \{chenjingjing,ygj\}@fudan.edu.cn\\
\{binzhu,cwngo\}@smu.edu.sg,\ $^{\ddag}$\emph{Corresponding author}
\end{tabular}
}
\begin{document}
\ninept
\maketitle
\begin{abstract}
Recent advances in Multimodal Large Language Models (MLMMs) have enabled recipe generation from food images, yet outputs often contain semantically incorrect actions or ingredients despite high lexical scores (e.g., BLEU, ROUGE). To address this gap, we propose a semantically grounded framework that predicts and validates actions and ingredients as internal context for instruction generation. Our two-stage pipeline combines supervised fine-tuning (SFT) with reinforcement fine-tuning (RFT): SFT builds foundational accuracy using an Action-Reasoning dataset and ingredient corpus, while RFT employs frequency-aware rewards to improve long-tail action prediction and ingredient generalization. A Semantic Confidence Scoring and Rectification (SCSR) module further filters and corrects predictions. Experiments on Recipe1M show state-of-the-art performance and markedly improved semantic fidelity.
\end{abstract}
\vspace{-1em}
\section{Introduction}
\label{sec:intro}

\vspace{-1em}
Recipe generation has become a key task in food computing, driven by health concerns and public datasets \cite{li2024cheffusion, recipe1m, dailyfood, gui2024navigating}. Early methods employed Transformer-based pipelines \cite{inversecooking, fire} that predicted ingredients before generating instructions. Large Multimodal Models (LMMs) such as FoodLMM \cite{foodlmm} enabled end-to-end generation, while Retrieval-Augmented Generation methods like RARG \cite{rarg} improved reliability by grounding outputs in retrieved knowledge. Retrieval-Augmented Generation (RAG) conditions generation on retrieved external evidence to improve factuality across NLP and vision (e.g., EVCAP, EXTRA) \cite{rag-survey,rag-relatedwork,cv_rag1,Evcap,EXTRA}; by contrast, we \emph{predict-as-retrieve} high-precision internal semantics (actions, ingredients) to ground generation.

\begin{figure}[t]
\centering
\includegraphics[width=1.0\columnwidth]{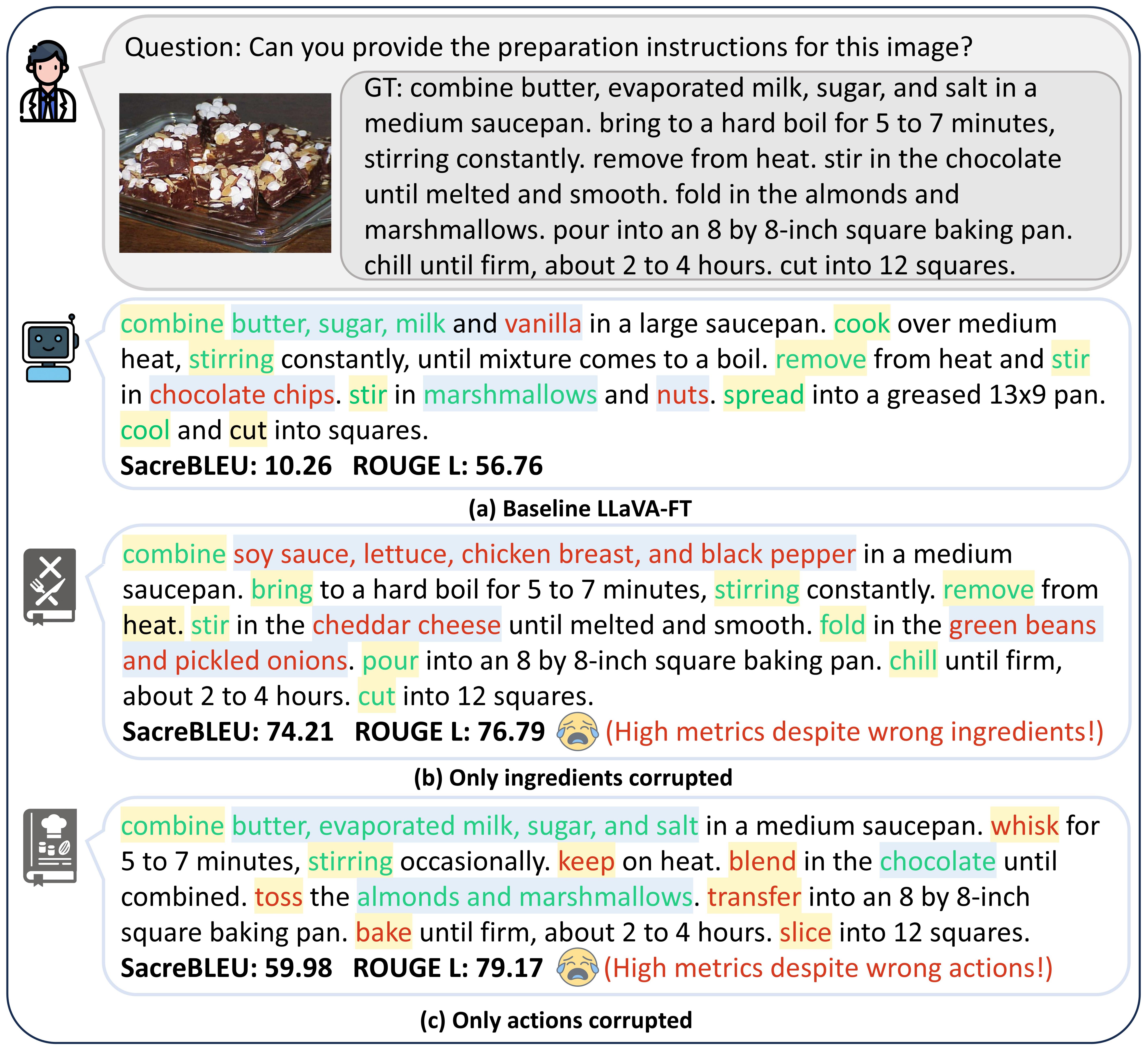}
\caption{Limitations of lexical metrics (e.g., SacreBLEU, ROUGE-L) in recipe generation. Comparison of LLaVA-FT output (a) with two semantically corrupted instructions: (b) correct actions but wrong ingredients, (c) correct ingredients but wrong actions. Red/green: incorrect/correct; yellow: actions; blue: ingredients.}
\vspace{-2em}
\label{sacrebleu-failure}
\end{figure}

\noindent Despite progress, models generate fluent but semantically incorrect recipes, e.g., with irrelevant ingredients or implausible actions \cite{inversecooking, fire}. Such errors are overlooked by standard metrics like SacreBLEU and ROUGE-L, which measure lexical overlap but fail to capture semantic fidelity (Figure \ref{sacrebleu-failure}). We argue that recipe integrity hinges on two core components—cooking actions and ingredients—and errors in these undermine coherence, reliability, and safety.

\noindent To address this, we propose a semantically grounded framework that explicitly predicts and validates actions and ingredients before instruction generation, analogous to RAG but without external retrieval. Our two-stage pipeline first employs Supervised Fine-Tuning (SFT) to establish baseline accuracy, then Reinforcement Fine-Tuning (RFT) with frequency-aware rewards to handle long-tail distributions. For action prediction, we introduce the Action Reasoning (AR) framework, combining a new Action-Reasoning dataset with Chain-of-Thought annotations~\cite{gpt4o} and Group Relative Policy Optimization (GRPO)~\cite{GRPO}. Ingredient prediction follows a similar pipeline, while a Semantic Confidence Scoring and Rectification (SCSR) module further enhances reliability.

\noindent Finally, we evaluate lexical fluency and semantic fidelity by complementing standard metrics with F1, IoU, recall, and precision computed on predicted actions and ingredients, demonstrating that our framework yields more accurate and semantically aligned recipes.
\begin{figure*}[t]
\centering
\includegraphics[width=1.0\textwidth]{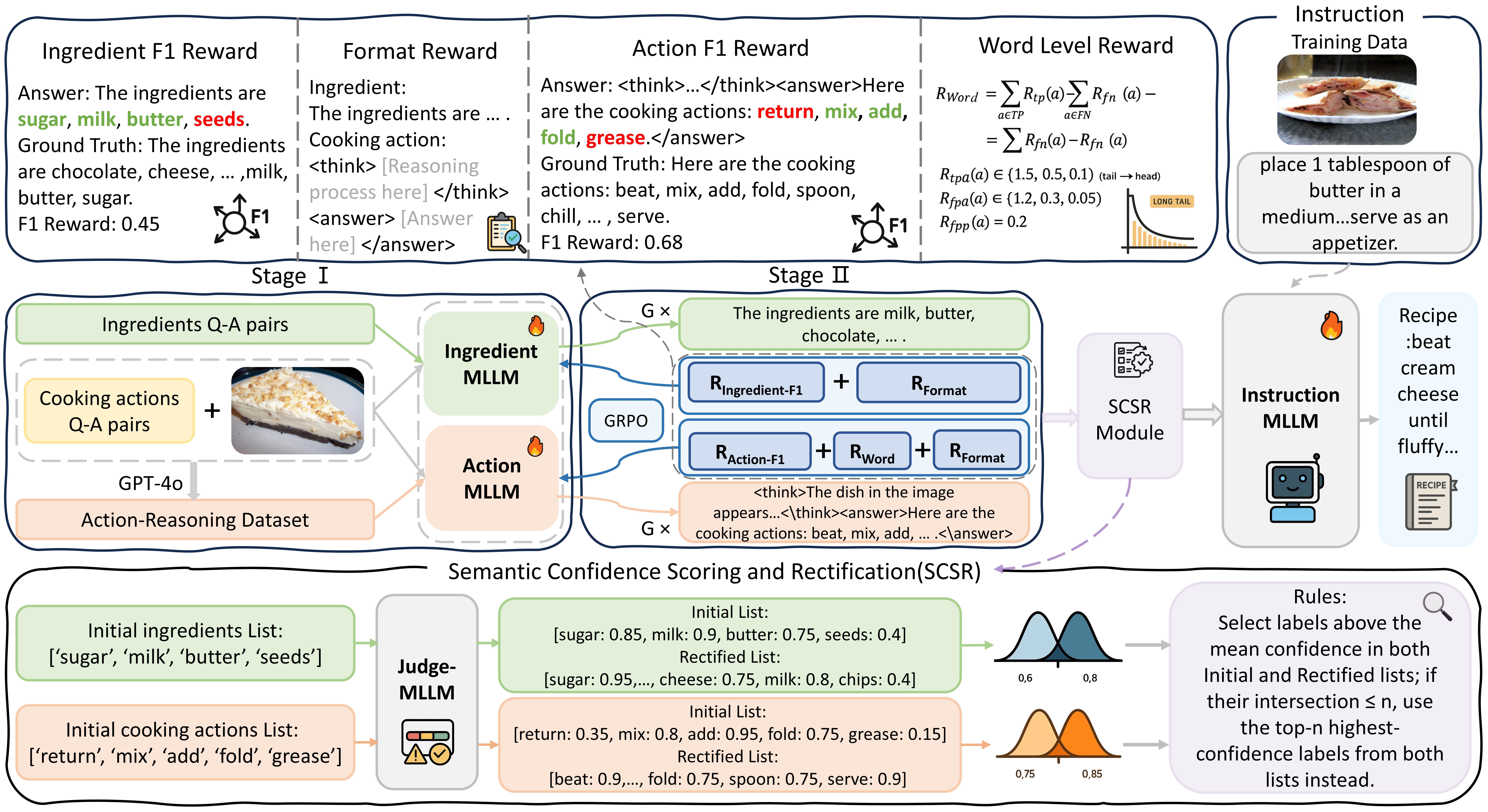} 
\caption{Overview of our \textbf{Semantically-Grounded Recipe Generation Framework}. The method prioritizes semantic integrity by modeling and verifying \textbf{ingredients} and \textbf{cooking actions} before instruction generation. A two-stage pipeline (SFT→GRPO) trains action/ingredient models; actions use CoT (AR-SFT) and frequency-aware rewards (AR-RFT) for long-tail. SCSR (MLLM, e.g., GPT-4o) scores/rectifies predictions; the generator conditions on the rectified labels.
}
\label{framework}
\vspace{-1.5em}
\end{figure*}
\vspace{-1.5em}
\section{Method}
\vspace{-1em}
% Our goal is to generate semantically faithful recipes from food images by explicitly modeling the core elements that define recipes: actions and ingredients. We introduce a two-stage generation framework. First, separate models predict cooking actions and ingredients from food images, trained via supervised and reinforcement fine-tuning, with a frequency-aware reward for long-tail actions. A Semantic Confidence Scoring and Rectification (SCSR) module filters low-confidence predictions to enhance label reliability before generation. These validated components are then prepended as structured context to guide instruction generation, improving factuality and semantic alignment. 
We generate semantically faithful recipes from images by explicitly modeling two core elements—actions and ingredients. Our two-stage framework trains separate predictors via SFT then RFT, with a frequency-aware reward to handle long-tail actions. A Semantic Confidence Scoring and Rectification (SCSR) module filters/corrects low-confidence labels. The validated context is prepended to the generator, improving factuality and semantic alignment.
\vspace{-0.8em}
\subsection{Two-Stage Training for Cooking Action Prediction}
\textbf{Multimodal Action Reasoning Data Construction}
% Directly fine-tuning Multimodal Large Language Models (MLLMs) for cooking action prediction is challenging due to the severe long-tail distribution of action labels. This imbalance encourages the model to default to predicting high-frequency actions to maximize accuracy, bypassing genuine visual reasoning. To address this and enhance the model's reasoning capability, we construct a multimodal Chain-of-Thought (CoT) dataset, referred to as the \textbf{Action-Reasoning} dataset, inspired by DeepSeek-R1-zero~\cite{Deepseek-r1}. The effectiveness of this Chain-of-Thought (CoT) approach over direct fine-tuning is validated in our ablation study (see Appendix), where models trained with CoT demonstrate improved accuracy and better generalization, especially on complex or long-tail action classes. To create this dataset, we first extract a vocabulary of the 383 most frequent action verbs from Recipe1M~\cite{recipe1m}. To further mitigate frequency bias, we filter out samples composed solely of the most common actions. The remaining data is then paired with food images and used as input prompts to GPT-4o \cite{gpt4o}, which generates detailed reasoning traces followed by the associated action labels. The design of the CoT prompt and a representative sample are shown in the Appendix.
Directly fine-tuning MLLMs for action prediction is hindered by a severe long-tail label distribution, leading models to overpredict frequent actions and bypass genuine visual reasoning. We therefore build a multimodal Chain-of-Thought \textbf{Action-Reasoning} dataset (inspired by DeepSeek-R1-zero \cite{Deepseek-r1}) to enforce explicit reasoning and boost tail sensitivity. Concretely, we extract 383 frequent action verbs from Recipe1M \cite{recipe1m}, filter samples dominated by very common actions, pair the remainder with images, and prompt GPT-4o \cite{gpt4o} to generate step-wise reasoning traces plus action labels. Training on this CoT corpus yields higher accuracy and better generalization, particularly for complex and long-tail classes.

\noindent \textbf{Stage 1: Action Reasoning Aware Supervised Fine-Tuning (AR-SFT)}
We fine-tune a pretrained MLLM (Qwen2.5-VL-7B) on the Action-Reasoning data with the standard SFT objective:
\begin{equation}
\mathcal{L}_{\text{SFT}}(\theta)=
\mathbb{E}_{(q,o)\sim\mathcal{D}}
\!\left[\frac{1}{|o|}\sum_{t=1}^{|o|}\log \pi_{\theta}(o_t\mid q,o_{<t})\right].
\end{equation}
Here, \(q\) is the image-conditioned prompt, \(o=(o_1,\dots,o_{|o|})\) the action sequence, and \(\pi_{\theta}\) the policy. This maximizes token likelihood given input and history, but SFT alone still biases the model toward frequent actions on ambiguous cases.

\noindent \textbf{Stage 2: Action Reasoning Reinforcement Fine-Tuning (AR-RFT)}
To address this limitation, we perform \textbf{Action Reasoning Reinforcement Fine-Tuning (AR-RFT)} using GRPO~\cite{GRPO}. For each query \(q\), the frozen old policy \(\pi_{\theta_{\text{old}}}\) samples \(G\) candidates \(\{o^{(i)}\}\) with rewards \(r^{(i)}\). We use normalized advantages
\begin{equation}
\hat{A}^{(i)}=\frac{r^{(i)}-\bar r}{\sigma_r}.
\end{equation}
The policy \(\pi_\theta\) is updated with a clipped ratio loss and a KL penalty to a reference policy (details omitted), effectively maximizing expected reward while remaining close to \(\pi_{\mathrm{ref}}\):
\begin{equation}
\max_{\theta}\ \mathbb{E}\!\left[R_{\text{Action-F1}}\right]-\beta_{\mathrm{KL}}\mathrm{KL}\!\big(\pi_\theta\| \pi_{\mathrm{ref}}\big).
\end{equation}
We adopt an F1-based multi-label reward to promote accuracy and completeness:
\begin{equation}
R_{\text{Action-F1}}=\frac{2\cdot \text{Precision}\cdot \text{Recall}}{\text{Precision}+\text{Recall}+\epsilon}.
\end{equation}
\begin{table*}[t]
\centering
\setlength{\tabcolsep}{2pt}
\scalebox{0.85}{
\begin{tabular}{l|ccc|cccc|cccc}
\hline
\multirow{2}{*}{Methods} & \multicolumn{3}{c|}{\textbf{Recipe Metrics}}               & \multicolumn{4}{c|}{\textbf{Action Metrics}}                           & \multicolumn{4}{c}{\textbf{Ingredient Metrics}}                        \\ \cline{2-12} 
                         & \textbf{SacreBLEU} & \textbf{ROUGE L} & \textbf{BERTScore} & \textbf{F1}    & \textbf{IOU}   & \textbf{Recall} & \textbf{Precision} & \textbf{F1}    & \textbf{IOU}   & \textbf{Recall} & \textbf{Precision} \\ \hline
InverseCooking \cite{inversecooking}  & 4.33 & 29.69 & 67.13 & 31.76 & 20.53 & 29.16 & 39.05 & 38.98 & 24.21 & 33.35 & 46.92 \\
FIRE \cite{fire}                       & 5.81 & 28.24 & 66.24 & 30.39 & 19.75 & 27.95 & 37.38 & 29.82 & 18.06 & 25.80 & 30.21 \\
FoodLMM \cite{foodlmm}                 & 5.82 & 35.09 & 67.93 & 32.62 & 22.13 & 30.79 & 40.07 & 31.19 & 18.48 & 26.20 & 38.54 \\
RARG \cite{rarg}                       & 6.01 & 38.51 & 68.26 & 36.75 & 24.36 & 34.95 & 42.89 & 37.20 & 22.85 & 33.11 & 42.46 \\
Qwen2.5-VL-7B‑FT \cite{qwenvl}                  & 8.34 & 36.83 & 69.22 & 40.61 & 27.31 & 41.40 & 43.43 & 43.24 & 27.58 & 39.10 & 48.34 \\
LLaVA-v1.5-7b‑FT \cite{llava-v1.5}             & \textbf{8.40} & 38.02 & 69.24 & 39.59 & 26.56 & 38.14 & 44.91 & 41.46 & 26.15 & 36.67 & 47.69 \\ \hline
Ours(Qwen2.5-VL-7B)          & 7.66 & 38.28 & 69.88 & 42.02 & 28.54 & 42.49 & 45.06 & 45.57 & 29.51 & 39.54 & 53.77 \\
\textbf{Ours(LLaVA-v1.5-7b)}& 8.08 & \textbf{41.08} & \textbf{69.98} & \textbf{42.35} & \textbf{29.82} & \textbf{40.03} & \textbf{49.21} & \textbf{47.24} & \textbf{30.93} & \textbf{43.75} & \textbf{51.34} \\ \hline
\end{tabular}
}
\caption{Performance comparison of recipe generation. Action and ingredient labels are extracted from the generated instructions for semantic evaluation. Best results for open‑source models are shown in \textbf{bold}.}
\label{main_table}
\vspace{-0.5em}
\end{table*}
Using only \(R_{\text{Action-F1}}\) in GRPO biases the policy toward frequent actions under the long-tail distribution. 
We therefore add a \textbf{Word-Level Reward} that weights TPs/FPs/FNs by label frequency to encourage rare but salient actions, with a fixed false-positive penalty \(R_{\text{fp}}(a)=0.2\):
\begin{equation}
R_{\text{Word}}=\sum_{a\in \text{TP}} R_{\text{tp}}(a)
-\sum_{a\in \text{FP}} R_{\text{fp}}(a)
-\sum_{a\in \text{FN}} R_{\text{fn}}(a).
\end{equation}
\vspace{-0.5em}
\begin{align}
R_{\text{tp}}(a) &=
\begin{cases}
1.5, & a \in \text{tail} \\
0.5, & a \in \text{mid} \\
0.1, & a \in \text{head}
\end{cases}
\qquad
R_{\text{fn}}(a) &=
\begin{cases}
1.2, & a \in \text{tail} \\
0.3, & a \in \text{mid} \\
0.05, & a \in \text{head}
\end{cases}
\end{align}
\vspace{-0.5em}
We further enforce structured outputs via a \textbf{Format Reward}:
\begin{equation}
\label{format}
R_{\text{Format}}=
\begin{cases}
1,& \text{if output matches the required pattern}\\
0,& \text{otherwise.}
\end{cases}
\end{equation}
\begin{itemize}
  \item \textbf{Pattern: }
  \texttt{\detokenize{^<think>.*?</think>}}\allowbreak
  \texttt{\detokenize{\s*<answer>.*?}}\allowbreak
  \texttt{\detokenize{</answer>}}
\end{itemize}

The overall action reward is
\begin{equation}
\label{reward-action}
R_{\text{action}}=\alpha\,R_{\text{Action-F1}}+\beta\,R_{\text{Format}}+\gamma\,R_{\text{Word}},
\end{equation}
where \(\alpha,\beta,\gamma\) weight each component.
\begin{table}[t]
\centering
\scalebox{0.85}{  % 根据需要调整缩放比例
\begin{tabular}{l|cccc}
\hline
Methods & F1 & IOU & Recall & Precision \\ \hline
InverseCooking \cite{inversecooking} & 48.44 & 31.96 & 48.23 & 48.66 \\
FIRE \cite{fire}                     & 49.19 & 32.61 & 49.17 & 49.21 \\
RARG \cite{rarg}                     & 49.78 & 33.14 & 49.86 & 49.70 \\
LLaVA-SFT \cite{llava-v1.5}          & 51.61 & 34.78 & 55.43 & 48.28 \\
Qwen-SFT \cite{qwenvl}               & 52.32 & 35.43 & \textbf{56.01} & 49.09 \\ \hline
\textbf{Ours}                        & \textbf{53.84} & \textbf{36.84} & 55.62 & \textbf{52.17} \\ \hline
\end{tabular}
}
\caption{Comparison of ingredient recognition results. ``LLaVA-SFT" and ``Qwen-SFT" denote LLaVA-v1.5-7B and Qwen2.5-VL-7B fine-tuned on Recipe1M ingredients; \textbf{Ours} is our ingredient model trained with SFT followed by RFT.}
\label{ingre_table}
\vspace{-1em}
\end{table}
\begin{figure*}[t]
\centering
\includegraphics[width=1.0\textwidth]{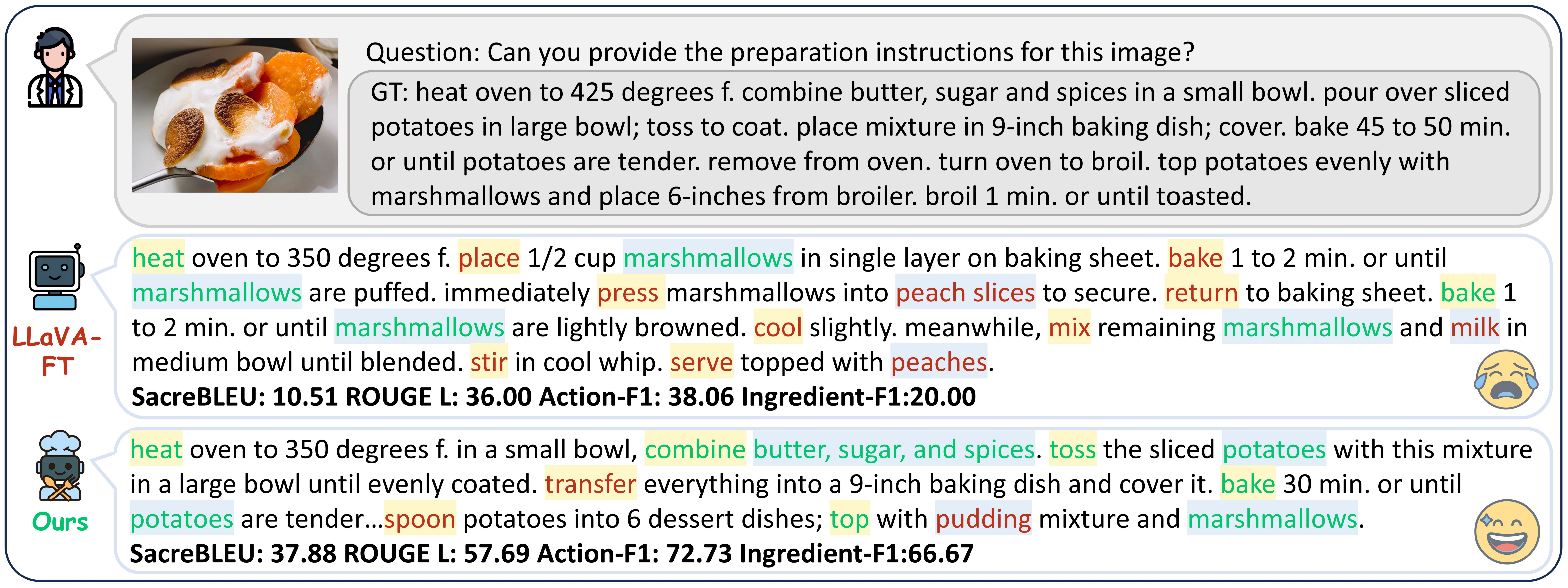} 
\caption{Qualitative Results. `GT' indicates the ground truth. Green text highlights correctly predicted cooking actions, ingredients, or phrases, while red text indicates incorrect predictions. Our model generates more complete, semantically aligned recipes than LLaVA-FT.}
\label{Qualitative}
\vspace{-1em}
\end{figure*}
% \begin{table}[h]
% \centering
% \begin{tabular}{l|cccc}
% \hline
% Methods  & F1             & IOU            & Recall         & Precision      \\ \hline
% Qwen-SFT & 41.09          & 27.73          & 38.56          & 48.71          \\
% \textbf{Ours}     & \textbf{41.88} & \textbf{28.65} & \textbf{39.47} & \textbf{48.96} \\ \hline
% \end{tabular}
% \vspace{-1em}
% \caption{Comparison of action prediction results. \textbf{Ours} represents the results from the LLaVA-v1.5-7B model, trained with AR-SFT followed by AR-RFT.}
% \label{main_action}
% \end{table}
\begin{table}[h]
\centering
\small
\setlength{\tabcolsep}{4pt}
\begin{tabular}{l|cccc}
\hline
Methods                    & F1             & IOU            & Recall         & Precision      \\ \hline
Ours                       & \textbf{41.88} & \textbf{28.65} & \textbf{39.47} & \textbf{48.96} \\
\hspace{2mm}-F1-Reward         & 39.51          & 25.78          & 38.87          & 46.24          \\
\hspace{2mm}-Word-Level Reward & 36.71          & 23.69          & 31.99          & 38.43          \\ \hline
\end{tabular}
\caption{Comparison of our action reward components, evaluated using the LLaVA-v1.5-7B model as the base.}
\label{ablation_action}
\vspace{-1em}
\end{table}
\begin{table*}[t]
\centering
\scalebox{0.85}{
\setlength{\tabcolsep}{2pt}
\begin{tabular}{l|c c c|c c c c|c c c c}
\hline
\multirow{2}{*}{Methods} & 
\multicolumn{3}{c|}{\textbf{Recipe Metrics}} & 
\multicolumn{4}{c|}{\textbf{Action Metrics}} & 
\multicolumn{4}{c}{\textbf{Ingredient Metrics}} \\ \cline{2-12}
& \textbf{SacreBLEU} 
& \textbf{ROUGE L} 
& \textbf{BERTScore} 
& \textbf{F1} 
& \textbf{IOU} 
& \textbf{Recall} 
& \textbf{Precision} 
& \textbf{F1} 
& \textbf{IOU} 
& \textbf{Recall} 
& \textbf{Precision} \\ \hline

LLaVA-FT & \textbf{8.40} & 38.02 & 69.24 & 39.59 & 26.56 & 38.14 & 44.91 & 41.46 & 26.15 & 36.67 & 47.69 \\ \hline

\quad +Action (AR-RFT) & 8.27 & 39.71 & 68.98 & 41.88 & 28.65 & 39.47 & 48.96 & 40.10 & 25.08 & 34.38 & 48.08 \\
\quad +Action (AR-RFT+SCSR) & 8.35 & 39.88 & 68.88 & 42.14 & 29.57 & 39.80 & 49.05 & 40.48 & 25.37 & 34.92 & 48.20 \\ \hline

\quad +Ingredient (RFT) & 7.78 & 39.43 & 68.96 & 38.92 & 26.23 & 35.87 & 46.85 & 45.00 & 29.03 & 41.07 & 49.77 \\
\quad +Ingredient (RFT+SCSR) & 7.84 & 41.05 & 68.96 & 38.97 & 26.38 & 35.98 & 46.78 & 46.10 & 29.96 & 41.96 & 51.16 \\ \hline

\quad \begin{tabular}[c]{@{}l@{}}+Action\&Ingredient\\ (AR-RFT+SCSR)\end{tabular} 
& 8.08 & \textbf{41.08} & \textbf{69.98} & \textbf{42.35} & \textbf{29.82} & \textbf{40.03} & \textbf{49.21} & \textbf{47.24} & \textbf{30.93} & \textbf{43.75} & \textbf{51.34} \\ \hline

\end{tabular}
}
\caption{Performance comparison of instruction generation under different action-ingredient context settings, highlighting the effect of the SCSR module. Metrics in the left block reflect lexical overlap (e.g., SacreBLEU, ROUGE-L), while those in the middle and right blocks reflect semantic alignment (e.g., BERTScore, F1, IOU).}
\label{ablation_SCSR}
\vspace{-1.5em}
\end{table*}
\vspace{-1em}
\subsection{Two-Stage Training for Ingredient Recognition}
% Different from cooking actions, ingredient recognition is a direct visual task that does not require complex reasoning, as ingredients can often be directly inferred from the image itself. We first construct the ingredient recognition data using the Recipe1M training set, where ingredient labels are formatted in the prompt as: ``The ingredients are ingredient$_1$, ingredient$_2$, $\ldots$''. Following a similar two-stage training strategy as used for actions, we first perform supervised fine-tuning on the constructed dataset, and then apply reinforcement fine-tuning using the GRPO training scheme to further improve model generalization. The reward function incorporates a multi-label F1 score and a format constraint that enforces the output to strictly follow the same pattern used in action prediction, defined as $\text{R}_{\text{format}}$ (see Eq.~\ref{format}): ``The ingredients are ingredient$_1$, ingredient$_2$, $\ldots$''. The multi-label F1 reward is computed as follows:
% \begin{equation}
% \text{R}_{\text{Ingredient-F1}} = \frac{2 \cdot \text{Precision} \cdot \text{Recall}}{\text{Precision} + \text{Recall} + \epsilon} \text{,}
% \end{equation}
% \noindent where $\epsilon$ is a small constant to avoid division by zero.
Unlike actions, ingredient recognition is largely a direct visual task. We construct a Recipe1M-based dataset with a fixed prompt template and train an ingredient predictor via a two-stage scheme: supervised fine-tuning followed by GRPO-based reinforcement fine-tuning. The reward combines a multi-label F1 term and a format constraint (Eq.~\ref{format}) to enforce the template. The F1 reward is
\begin{equation}
R_{\text{Ingredient-F1}}
=\frac{2\cdot \text{Precision}\cdot \text{Recall}}
{\text{Precision}+\text{Recall}+\epsilon},
\end{equation}
where \(\epsilon\) avoids division by zero.
\vspace{-1em}
\subsection{Semantic Confidence Scoring and Rectification}
Predictions from previous stages still contain non-trivial false positives. We therefore add a \textbf{Semantic Confidence Scoring and Rectification (SCSR)} post-processor that uses a Judge-MLLM (GPT-4o) to rescore and refine labels.

Given initial predictions \(L_{\text{init}}=\{l_i\}_{i=1}^{k}\), SCSR returns
\(S_{\text{init}}=\{(l_i,c_i)\}_{i=1}^{k}\) and a rectified set
\(S_{\text{rect}}=\{(l'_j,c'_j)\}_{j=1}^{m}\). We define mean thresholds
\begin{equation}
\tau_{\text{before}}=\frac{1}{k}\sum_{i=1}^{k}c_i,\qquad
\tau_{\text{after}}=\frac{1}{m}\sum_{j=1}^{m}c'_j,
\end{equation}
keep high-confidence labels
\(L_{\text{before}}=\{l_i\,|\,c_i>\tau_{\text{before}}\}\),
\(L_{\text{after}}=\{l'_j\,|\,c'_j>\tau_{\text{after}}\}\),
and take \(L_{\text{inter}}=L_{\text{before}}\cap L_{\text{after}}\).
To guarantee sufficient context, we use a top-\(n\) fallback:
\begin{equation}
\label{topn}
L_{\text{final}}=
\begin{cases}
L_{\text{inter}}, & |L_{\text{inter}}|\ge n,\\
\text{Top-}n\!\big(S_{\text{init}}\cup S_{\text{rect}}\big), & \text{otherwise},
\end{cases}
\end{equation}
where \(\text{Top-}n(S)\) selects the \(n\) unique labels with highest confidence.
This balances precision (via intersection) and recall (via top-\(n\)), yielding more reliable context for final generation.
\vspace{-1em}
\subsection{Instruction Generation with Action-Ingredient Context Prompting}
% Inspired by retrieval-augmented generation (RAG), we adopt a modular inference strategy for instruction generation. Specifically, the predicted cooking actions and ingredients from preceding modules are \textit{only concatenated at inference time} as contextual cues, without joint training with the instruction generation model. This preserves modularity while leveraging task-specific semantic information. We construct the input prompt in the following format:
% \begin{quote}
% \texttt{Here are the cooking actions: [...]. The ingredients are [...]. Can you provide the preparation instructions for this image?}
% \end{quote}

% By prepending these structured predictions, the model is guided to attend to relevant content, improving both coherence and factual correctness in the generated instructions. Because no joint tuning is needed, our prompt can be easily applied to any fine-tuned multimodal model, making the approach broadly compatible and easy to use.
Inspired by retrieval-augmented generation (RAG), we adopt a modular inference strategy: predicted \textit{actions} and \textit{ingredients} are concatenated only at inference as structured context (no joint training), preserving modularity while injecting task-specific semantics. We use the prompt:
\texttt{Here are the cooking actions: [...]. The ingredients are [...]. Can you provide the preparation instructions for this image?}
By prepending these structured predictions, the model is guided to attend to relevant content, improving both coherence and factual correctness in the generated instructions. Because no joint tuning is needed, our prompt can be easily applied to any fine-tuned multimodal model, making the approach broadly compatible and easy to use.
\vspace{-1.5em}
\section{Experiments and Results}
\vspace{-1em}
\subsection{Implementation Details}
\vspace{-0.5em}
All experiments use four NVIDIA A100 80GB GPUs with \texttt{bf16}. 
For both the \emph{cooking-action} and \emph{ingredient} recognizers, we fine-tune \textit{Qwen2.5-VL-7B}~\cite{qwenvl} with LoRA~\cite{lora} (rank 128, $\alpha{=}32$) in two stages:
(i) SFT for 1 epoch (AdamW, lr $2{\times}10^{-5}$, batch 8, context 2048, 5\% warm-up);
(ii) GRPO~\cite{GRPO} RFT with lr $5{\times}10^{-6}$, effective batch 64 (gradient accumulation), and $G{=}8$ candidates per query using the composite reward (\textsc{F1}+word-level+format).
The \emph{instruction generator} uses two backbones:
(i) \textit{Qwen2.5-VL-7B} with the same LoRA/schedule as above, context 4096;
(ii) \textit{LLaVA-v1.5-7B}~\cite{llava-v1.5} with LoRA rank 128, lr $2{\times}10^{-5}$, batch 8, context 4096 (public visual-instruction-tuning recipe).
Decoding: temperature 0.2, nucleus $p{=}0.9$, top-$k{=}40$.
All baselines are re-evaluated under the same settings and metrics for fairness.
% 4×A100-80GB, \texttt{bf16}. Recognizers (actions/ingredients): Qwen2.5-VL-7B+LoRA (rank 128, $\alpha{=}32$). SFT (1 epoch, AdamW, lr $2{\times}10^{-5}$, batch 8, ctx 2048, 5\% warm-up) → GRPO RFT (lr $5{\times}10^{-6}$, eff. batch 64 via accumulation, $G{=}8$, reward=\textsc{F1}+word+format). Instruction generator: (i) Qwen2.5-VL-7B (same LoRA/schedule, ctx 4096); (ii) LLaVA-v1.5-7B (rank 128, lr $2{\times}10^{-5}$, batch 8, ctx 4096). Decoding: temp 0.2, nucleus $p{=}0.9$, top-$k{=}40$. All baselines re-evaluated under the same settings.
\vspace{-1.5em}
\subsection{Datasets and Metrics}
\vspace{-0.5em}
\textbf{Datasets.} We use Recipe1M~\cite{recipe1m} (title, ingredients, stepwise instructions); ingredients follow 1{,}488 classes~\cite{inversecooking}. One image per recipe yields 249{,}603 train pairs for SFT (ingredients) and supervised instruction training; 30k samples are used for ingredient RFT. Cooking actions are extracted with Llama-3.1~\cite{llama3}; 30k CoT samples are generated by GPT-4o~\cite{gpt4o} and filtered by confidence $>{}0.85$. To reduce frequency bias, we drop samples whose actions all fall in the top-55 verbs, leaving 10{,}397 for AR-SFT; for AR-RFT, 6k extra samples are filtered similarly, yielding 3{,}712. Evaluation uses a fixed 1{,}444-sample subset.

\noindent\textbf{Metrics.} Ingredients: F1, IoU. Recipes: SacreBLEU, ROUGE-L, BERTScore~\cite{bertscore}. For semantic fidelity, we extract actions/ingredients from generated text and report F1/IoU/precision/recall against ground truth.
\vspace{-1em}
\subsection{Performance Comparison}
\vspace{-0.5em}
\textbf{Quantitative Comparison of Recipe Generation.}To demonstrate the general applicability of our framework, we integrated our predicted action and ingredient context into two distinct MLLMs: Qwen2.5-VL-7B and LLaVA-v1.5-7B. 
% As shown in Table~\ref{main_table}, although this results in a slight decrease in the lexical-based SacreBLEU score, in contrast, our method achieves substantial improvements on semantic metrics, including BERTScore, Action-F1, and Ingredient-F1, which evaluate contextual alignment and the correctness of key structured information. This trade-off aligns with our goal of prioritizing semantic accuracy over superficial lexical similarity. As the LLaVA-FT model exhibited stronger overall performance with our enhancements, we selected it as the base for all subsequent ablation studies.
As Table~\ref{main_table} shows, while SacreBLEU drops slightly, semantic metrics (BERTScore, Action-F1, Ingredient-F1) improve markedly, indicating better contextual alignment and correctness of key structure. Given the stronger overall gains with our enhancements, we use LLaVA-FT as the base for subsequent ablations.
% \vspace{-1em}

\noindent\textbf{Quantitative Comparison of Ingredients Recognition.}
As shown in Table~\ref{ingre_table}, we compared our two-stage trained ingredient model, built upon Qwen2.5-VL-7B, with other state-of-the-art methods. Our model achieves the best overall performance, underperforming the Qwen-SFT baseline only slightly on the recall metric, which favors precision by excluding low-confidence labels.
% Table~\ref{ingre_table} shows our two-stage ingredient model (Qwen2.5-VL-7B) achieves the best overall scores; it trails Qwen-SFT slightly on recall due to stricter precision.
\vspace{-1em}
\subsection{Ablation Study}
\textbf{Ablation of AR-RFT Rewards in Cooking Action Prediction.}
% We present a comparison of cooking action model performance after AR-SFT and AR-RFT in Table~\ref{main_action}. We also perform an ablation study to assess the impact of F1 and Word-Level rewards in the second-stage GRPO training for cooking action prediction (Table~\ref{ablation_action}). Since the SFT stage already enforces well-structured outputs, all variants include the \textbf{Format Reward} by default. We fix the weights in the reward function (Eq.~\ref{reward-action}) as $\alpha = 0.1$, $\beta = 1$, and $\gamma = 1$. Using only the F1 Reward leads to high precision but low recall due to conservative predictions. In contrast, the Word-Level Reward encourages rare action prediction, increasing recall but introducing more false positives. Combining both yields the best balance between accuracy and coverage. 
We ablate the GRPO rewards (Table~\ref{ablation_action}). All variants include the \textbf{Format Reward}; Eq.~\ref{reward-action} weights are fixed to \(\alpha{=}0.1,\ \beta{=}1,\ \gamma{=}1\). Using only the F1 reward yields conservative predictions (high precision, low recall); using only the Word-Level reward boosts recall by promoting rare actions but increases false positives. Combining both achieves the best balance (F1/IoU).
% \vspace{-0.5em}

\noindent\textbf{Ablation Study on Action-Ingredient Prompting and SCSR.}
% We conduct a hybrid ablation study to validate two key aspects of our framework: the efficacy of the Semantic Confidence Scoring and Rectification (SCSR) module and the impact of using predicted cooking actions and ingredients as supplementary context for recipe generation. As detailed in Table~\ref{ablation_SCSR}, we assess the final instruction model under several conditions. We provide the context of actions alone, ingredients alone, and both combined, comparing the results of using the direct outputs from the GRPO-trained models versus using the outputs purified by our SCSR module. The SCSR module uses Multimodal LLMs (MLLMs), including \textbf{GPT-4o} and \textbf{Qwen2.5-VL-72B} \cite{qwen2.5-VL}, to revise predicted labels. Results in the main text are based on GPT-4o, while the Appendix provides additional experiments using Qwen2.5-VL-72B. The top-$n$ selection used in Equation~\ref{topn} takes $n=3$ throughout all settings. The results yield two clear conclusions. First, the application of the SCSR module leads to performance gains across nearly all metrics, confirming that it effectively improves the reliability of the predicted cooking actions and ingredients. Second, the results show that using both cooking actions and ingredients as a combined context is superior to using either individually, highlighting their synergistic effect in guiding the final recipe generation.
We run a hybrid ablation to evaluate (i) the SCSR module and (ii) the benefit of using predicted actions/ingredients as context (Table~\ref{ablation_SCSR}). We test actions only, ingredients only, and both, and for each compare raw GRPO outputs vs. SCSR-refined labels. SCSR uses an adjudicator MLLM (GPT-4o; similar trends with Qwen2.5-VL-72B~\cite{qwenvl}). The top-$n$ in Eq.~\ref{topn} is set to $n{=}3$. Results show: (1) SCSR improves nearly all metrics by producing more reliable labels; (2) combining actions+ingredients outperforms either alone, confirming their complementarity for recipe generation.
\vspace{-1em}
\subsection{Qualitative Results}
\vspace{-0.5em}
% Figure~\ref{Qualitative} presents a qualitative comparison between our proposed model, LLaVA-FT, and the ground truth (GT). As shown, our model demonstrates higher precision than LLaVA-FT in predicting ingredients, cooking actions, and other details like cooking tools, enabling it to generate more specific and accurate instructions. As illustrated of Figure~\ref{Qualitative}, our model achieves more accurate ingredient and action prediction compared to LLaVA-FT. It correctly identifies key ingredients such as \textit{butter}, \textit{sugar}, and \textit{spices}, while LLaVA-FT misses these and incorrectly predicts \textit{peaches}. For actions, our model captures essential steps like \textit{combine}, \textit{toss}, and \textit{bake}, and also aligns with the ground truth on cooking states such as \textit{until potatoes are tender}, which LLaVA-FT omits. This highlights our model’s stronger semantic grounding. This ability to correctly identify crucial semantic components directly contributes to our model's significantly higher \textbf{Action-F1} and \textbf{Ingredient-F1} scores.
Figure~\ref{Qualitative} compares our model with LLaVA-FT and ground truth (GT). Our model achieves higher precision on ingredients, actions, and cues (e.g., tools), producing more specific instructions. It correctly identifies \textit{butter}, \textit{sugar}, and \textit{spices}, while LLaVA-FT misses these and wrongly predicts \textit{peaches}. For actions, it captures \textit{combine}, \textit{toss}, and \textit{bake}, and matches GT on states (e.g., \textit{until potatoes are tender}) that LLaVA-FT omits. These gains explain the higher \textbf{Action-F1} and \textbf{Ingredient-F1} scores.
\vspace{-1.5em}
\section{Conclusion}
\vspace{-0.5em}
% In this work, we address the failure of lexical metrics to capture semantic errors in recipe generation by proposing a framework that first explicitly predicts and refines the core components—cooking actions and ingredients—to serve as a high-fidelity internal context. Our two-stage Action Reasoning (AR) training, which uses a Chain-of-Thought dataset and GRPO with frequency-aware rewards, along with our Semantic Confidence Scoring and Rectification (SCSR) module, leads to state-of-the-art performance on the Recipe1M benchmark for both ingredient recognition and overall recipe generation. This work paves the way for more semantically-grounded evaluation and generation strategies in multimodal food AI.
We tackle the limits of lexical metrics in recipe generation by predicting and refining core components—cooking actions and ingredients—as high-fidelity context. Our two-stage Action Reasoning training with Chain-of-Thought data, GRPO-based rewards, and a Semantic Confidence Scoring and Rectification (SCSR) module achieves SOTA results on Recipe1M, advancing semantically grounded recipe generation.
\vfill\pagebreak

\section{ACKNOWLEDGEMENT}
This research/project is supported by the Singapore Ministry of Education (MOE) Academic Research Fund (AcRF) Tier 1 grant (Proposal ID: 23-SIS-SMU-023) and Tier 2 (Proposal ID: T2EP20222-0046). Any opinions,
findings and conclusions or recommendations expressed
in this material are those of the authors and do not reflect the views of the Ministry of Education, Singapore.
% \label{sec:refs}

% List and number all bibliographical references at the end of the
% paper. The references can be numbered in alphabetic order or in
% order of appearance in the document. When referring to them in
% the text, type the corresponding reference number in square
% brackets as shown at the end of this sentence \cite{C2}. An
% additional final page (the fifth page, in most cases) is
% allowed, but must contain only references to the prior
% literature.

% Please follow the IEEE Citation Guidelines, \url{https://ieee-dataport.org/sites/default/files/analysis/27/IEEE\%20Citation\%20Guidelines.pdf} for formatting of references.

\bibliographystyle{IEEEbib}
\bibliography{strings,refs}

\end{document}